# Game-theoretic applications of a relational risk model

Tatiana Urazaeva[1][0000-0003-2649-8698]

[1] Volga State University of Technology, Lenin Sq. 3, 424000 Yoshkar-Ola, Russia
urazaevata@volgatech.net

**Abstract.** The report suggests the concept of risk, outlining two mathematical structures necessary for risk genesis: the set of outcomes and, in a general case, partial order of preference on it. It is shown that this minimum partial order should constitute the structure of a semilattice. In some cases, there should be a system of semilattices nested in a certain way. On this basis, the classification of risk theory tasks is given in the context of specialization of mathematical knowledge. In other words, we are talking about the development of a new relational risk theory.

The problem of political decision making in game-theoretic formulation in terms of having partial order of preference on the set of outcomes for each participant of the game forming a certain system of nested semilattices is considered as an example of a relational risk concept implementation. Solutions to the problem obtained through the use of various optimality principles are investigated.

## 1 Introduction

Nowadays, the concept of risk has firmly established itself in a large variety of economic, technological, political and other areas. The first scientific approaches aimed at the investigation of risk date back to the $18^{th}$ century to the works of the Swiss mathematician Daniel Bernoulli who studied the paradox of tossing a coin which was later called the St. Petersburg paradox [18]. Since then, the number of publications on the topic of risk has been booming. At the end of the $20^{th}$ century and at the beginning of the $21^{st}$ century, Ulrich Beck [1, 2, 17], Anthony Giddens [5] and Niklas Luhmann [10, 11] provided some versions of an integral concept of "risk society" and even of "world risk society" [1]. Public awareness of the significance of the risk concept has eventually led to the standardization processes in the field of risk management [6, 7, 8].

However, despite such a long evolution of the risk concept, the extent of the corresponding discourse within society, and the fact that risk management processes are now regarded on a commercial scale, a universal concept of risk has not been coined, a mathematically rigorous formal vision of risk has not been provided, the ideas concerning general properties of risk as a mathematical object have not been chosen either.

In this paper, we suggest an attempt to coin the most common concept of risk which is based on the concept of relation. In fact, it will cover the development of a relational risk theory. An important aspect of any theory is its predictive function implemented through scientific tools provided by the theory. To illustrate the use of the corresponding tools, we provide a decision making example in the area which, on the one hand, refers to the field of macroeconomics and, on the other hand, to the field of politics. In this regard, the search for the best solution is done through the game theoretic task setting of the risk management in the absence of quantitative evaluation of game outcomes. The task set in such a way is considered to be interesting because it makes the most of the approaches offered by a new risk theory.

## 2  Basic ideas of the risk concept

First of all, when establishing a relational risk theory, it is necessary to notice that risk is possible only if there is some development, where the future is diverse, $|\Omega| > 1$, see Fig. 1.

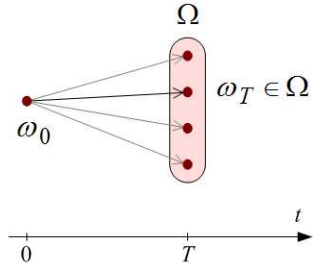

**Fig. 1.** System evolution from a state $\omega_0$ at the initial time into a state $\omega_T$ on the planning time-frame $T$, $\omega_T \in \Omega$.

Here, $\Omega$ is a set of possible outcomes of the system development. To simplify the considerations, we will suppose that this set $\Omega$ is finite.

We will describe a decision making procedure through a simple function of system response as follows:

$$f: \Delta \to \Omega$$

where $\Delta$ is a set of decision maker's actions. In this paper we do not intend to consider the nature of the function $f$, but we will focus on the decision maker's preferences which might occur on the set $\Omega$.

First, let us assume that the decision maker's preference is a preference of a (partial) order. Fig. 2 shows different variants of a (partial) order preference on the set $\Omega$ which can be spoken about in terms of system development risk. The preference is shown in the Fig. 1 through Hasse diagrams presented by thick arrows. The simplest case shown in Fig. 2 is a case of a total order (c). This situation is considered to be a standard one for the classical risk theory: all the outcomes are comparable; there is a risk of decision-making which does not lead to the best for a decision maker outcome. The best outcome here is $\omega_1$. Case (a) shown in the Figure represents a partial order which induces a structure of an upper semilattice on $\Omega$. This case was considered in the paper [16] and in monographs [14, 15]. There is also a decision-making risk which does not lead to the best for a decision maker outcome. In this regard, the best outsome here is also $\omega_1$.

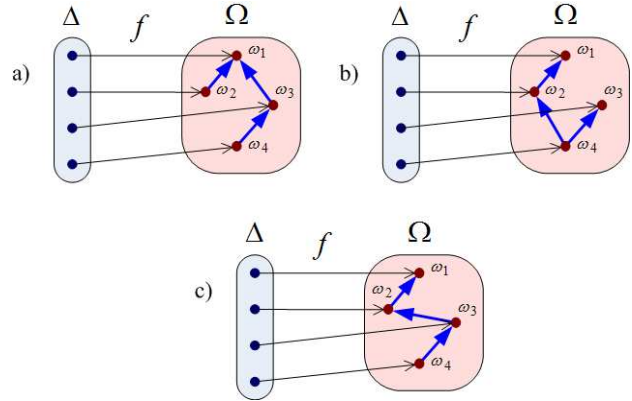

**Fig. 2.** A system response function and preference variants on the set $\Omega$ which correspond to the risk situation.

Case (b) is a new situation considered in the present paper for the first time. Here, a decision maker induces his/ her preference on the lower semilattice structure $\Omega$. This case should be interpreted as a task to avoid the risk of unfavorable outcomes, the worst of which is $\omega_4$. The best solutions in the case under consideration should be those which lead to the outcomes from the set of the maximum ones $\{\omega_1, \omega_3\}$.

Now, let us consider the cases of partial orders the presence of which does not allow us to conclude straightaway, without additional constructions, that there is some risk. The simplest example of this is case (a) shown in Fig. 3. Here, a decision maker does not have any preferences at all. In case (b) there is an outcome $\omega_4$ which cannot be compared with the others, whereas there is the same outcome $\omega_1$ in case (d). In case (c) there are two subsets $\{\omega_1, \omega_2\}$ and $\{\omega_3, \omega_4\}$ the outcomes of which cannot be compared with each other.

Taking into account the given cases, only case (c) can be most naturally considered in terms of risk situation. To do so, we shall introduce an indifference relation A in the following way: $\omega_1 \, A \, \omega_3$, $\omega_2 \, A \, \omega_4$. Then a decision maker induces his/her preference on factor set $\Omega / A$ as a total order with an obvious interpretation within risk theory. The best solution here is the choice of an action which leads to one of the outcomes of equivalence class $\omega_1 / A = \omega_3 / A = \{\omega_1, \omega_3\}$.

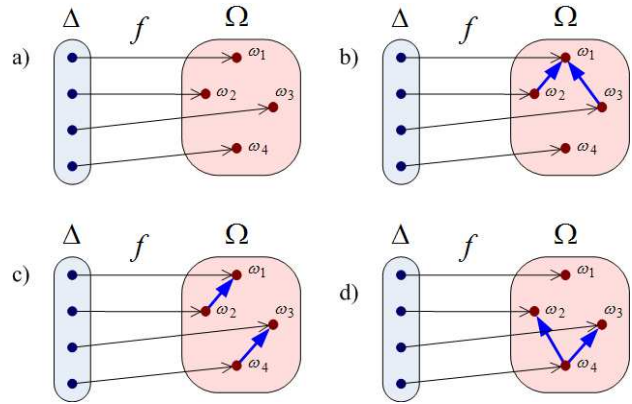

**Fig. 3.** System response function and preference variants on the set $\Omega$ which do not directly correspond to the risk situations

In cases (b) or (d) in Fig. 3, we can do the same thing introducing an outcome $\omega_4$ or $\omega_1$ respectively in the equivalence class to one of the outcomes left. This is the least obvious situation from the interpretation point of view but it is possible! In both cases, the preference will induce the structure of upper or lower semilattice on factor sets, respectively. Thus, it is possible to say that there is some risk.

Having considered the given examples, we can conclude that as a fundamental element of risk we should accept that there is a partial order on the set of the outcomes which has a structure of a semilattice (the upper one is when aiming at the best outcome, while the lower one is when trying to avoid the worst one). To provide the understanding of the degree of concept generalization, it should be noticed that a total order is a particular case of a semilattice (both upper and lower ones).

The used procedure, when we reduce partially ordered sets to the factor sets with semilattice structure, can be successfully applied for the investigation of preorder preferences to interpret situations in complex economic systems having a significant lack of information within risk theory.

To sum it up, it is necessary to notice that relational risk theory which connects risk and preference on the set of the outcomes forming the semilattice structure is considered to be the most common one. The present concept summarizes the results given both in the works covering the theoretical basics of information security [4, 9] and in the works referring to the decision-making with the sets having non-numeric elements [16].

## 3    Relational risk theory and solution concepts

While investigating the reference of the risk genesis in developing systems to solution concepts (and to the corresponding areas of mathematics), it is convenient to step back from the nature of the processes which take place inside them and focus on the role of one or several decision makers. Let us consider some several cases.

1. Let us assume that $\Delta$ is a set of the possible actions (decisions) of the only one decision maker in a system. Then, let $f : \Delta \to \Omega$ be the function which describes system development outcome $\omega \in \Omega$ as a result of making this decision $\delta \in \Delta$.

If $|\Omega| = 1$ no matter how powerful $\Delta$ is, then we have a strictly predefined development which is not affected by the intentions of a decision maker. Consequently, in this situation we cannot speak about any risk.

If $|\Omega| > 1$ and $\Omega$ together with the set on it preference partial order form the upper (disjunctive) semilattice [7], then there is such an optimal solution $\delta^*$ that $f(\delta^*) = \sup \Omega$. It should be noticed that the solution of such optimization problems is the task of mathematical programming. In this case, it is possible to speak about the risk of a nonoptimal choice. Technically, this kind of risk should be referred to as risk under conditions of (full) certainty.

Here, we assume that a risk under conditions of full certainty is a risk when a decision maker makes a mistake when optimizing the function *f* on the set of possible actions $\Delta$. The mistake can be caused by an ordinary lack of attention of a decision maker or it can result from a lack of computing resources if *f* is rather complex. In this case we can speak about full certainty as mathematically a decision maker possesses all the information necessary for decision-making. Later on, all the risks associated with the decision maker's mistakes or scarcity of computing resources under condition of full awareness will be referred to as risks under conditions of full certainty.

2. Let $N$ be a fixed community of decision makers acting within a given system and having contradictory or non-contradictory interests relative to each other. Then, let $\Delta_i$ be a set of possible actions (decisions) of an $i$-decision maker, $i \in N$, $f : \Pi_{i \in N} \Delta_i \to \Omega$ is a function which describes the outcome of the economic development $\omega \in \Omega$ as a result of decision $\delta_i \in \Delta_i$, $i \in N$ made by each agent.

As in the previous case, if $|\Omega| = 1$, then we will have a strictly predefined development which is not affected by the intentions of a decision maker. Thus, we cannot speak about any risk.

Let us assume that $|\Omega| > 1$. We shall consider the situation from the point of view of an $i$-decision maker. Let us introduce a notation to define the set of decisions made by an $i$-decision maker on the set of the outcomes of economic development if the choices made by the other community members are fixed: $f\left[\delta_{\hat{i}}\right](\delta_i) = f\left(\delta_{\hat{i}}, \delta_i\right)$, $\delta_i \in \Delta_i$, $\delta_{\hat{i}} \in \Pi_{j \in N \setminus \{i\}} \Delta_j$, $i \in N$. Besides, we use notation $\text{Cod}(g)$ to define the domain of a function $g$. If a partial preference order which defines the upper semilattices on $\Omega$ and on all its subsets of $\operatorname{Cod}\left(f\left[\delta_{\hat{i}}\right]\right)$ type, $\delta_{\hat{i}} \in \Pi_{j \in N \setminus \{i\}} \Delta_j$ is assigned only for an $i$-decision maker, then depending on the situation (or the setting) there occurs either a parametric problem of mathematical programming or a game under full uncertainty. Thus, we can speak about two types of risks: in some cases it is a risk of choice of unfavorable solutions for an $i$-decision maker by a community $N \setminus \{i\}$; and in all cases it is a risk of a nonoptimal choice of an $i$-decision maker himself. If the same partial preference order is defined at least for one more decision maker apart from an $i$-decision maker, then there occurs a problem of game theory in pure strategies. Here, apart from a risk of a nonoptimal choice (in any sense), in some cases we can also speak about a risk of an unfavorable choice for an $i$-decision maker provided by a community $N \setminus \{i\}$. Here, if the preference is defined for all decision makers, in some cases (if all the players behave rationally) these types of risks can be regarded as risks under (full) uncertainty.

3. Let us assume that, as in the first case, $\Delta$ is a set of possible actions of the only one decision maker in the system. Then, let $(\Omega_0, \mathcal{A}_0, P_0)$ be a probabilistic space describing the uncertainty of the environment (nature), $\Omega_0$ is a set of environment states, $\mathcal{A}_0$ is $\sigma$-algebra of subsets of the set $\Omega_0$, $P_0$ is a probability measure on the measurable space $(\Omega_0, \mathcal{A}_0)$, $f : \Delta \times \Omega_0 \to \Omega$ is a function describing outcome $\omega \in \Omega$ as a result of decision making $\delta \in \Delta$ at environmental state $\omega_0 \in \Omega_0$ given. Let us assume that there is also a $\sigma$-algebra of subsets $\mathcal{A}$ defined for a set of outcomes $\Omega$ ($|\Omega| > 1$). We will also assume that with every fixed $\delta \in \Delta$ the function $f_\delta : \Omega_0 \to \Omega$ which is defined by the ratio of $f_\delta(\omega_0) = f(\delta, \omega_0)$, $\omega_0 \in \Omega_0$ is measurable in relation to a pair of $\sigma$-algebras $\mathcal{A}_0$ and $\mathcal{A}$, i.e. $f_\delta^{-1}(A) \in \mathcal{A}_0$ is for arbitrary set $A \in \mathcal{A}$. It means that, among other factors, every solution $\delta \in \Delta$ causes probability measure $P_\delta$ on measurable space $(\Omega, \mathcal{A})$ according to the rule: $P_\delta(A) = P_0\left(f_\delta^{-1}(A)\right)$, $A \in \mathcal{A}$. Let us consider the set of measures $\mathcal{P} = \{P_\delta : \delta \in \Delta\}$. There is some risk even if $|\mathcal{P}| = 1$ as long as the same structure of preference in defined on the set $\Omega$. Here, the risk is associated with the randomness of an outcome. In this case, we have a risk of an unfavorable environment state.

If $|\mathcal{P}| > 1$, when $\mathcal{P}$ already has a minimum structure of upper semilattice, we can solve a problem of the game theory with the nature looking for such $\delta^*$, that $P_{\delta^*} = \sup \mathcal{P}$. Apart from risk of an unfavorable environment state, there can occur a risk of a nonoptimal choice.

4. Let us assume, as in the second case, that $N$ is a fixed community of decision makers acting within a given system. Then, let $\Delta_i$ be a set of the possible actions (decisions) of an $i$-decision maker, $i \in N$, $f : \Pi_{i \in N} \Delta_i \to \Omega$ is a function which describes the outcome of economic development $\omega \in \Omega$, as a result of a decision made by each decision maker $\delta_i \in \Delta_i$, $i \in N$. Let us assume that the choice of an action made by an $i$-decision maker is random and it can be described through probabilistic space $(\Delta_i, \mathcal{D}_i, P_i)$, where $\mathcal{D}_i$ is $\sigma$-algebra of subsets of set $\Delta_i$, $P_i$ is a probability measure on measurable space $(\Delta_i, \mathcal{D}_i)$, $i \in N$. In this case, the strategy of an $i$-decision maker is the probability measure $P_i$, $i \in N$. Then, let us assume that $(\Pi_{i \in N} \Delta_i, \otimes_{i \in N} \mathcal{D}_i, \Pi_{i \in N} P_i)$ is a product of probabilistic spaces $(\Delta_i, \mathcal{D}_i, P_i)$, $i \in N$, [13], and $\sigma$-algebra of subsets $\mathcal{A}$ is defined on the set of outcomes $\Omega$ ($|\Omega| > 1$). In this case, the function $f$ is measurable in relation to $\sigma$-algebra pair of $\otimes_{i \in N} \mathcal{D}_i$ and $\mathcal{A}$, i.e. $f^{-1}(A) \in \otimes_{i \in N} \mathcal{D}_i$ is for arbitrary set $A \in \mathcal{A}$. Let us define measure $P(A) = (\Pi_{i \in N} P_i)(f^{-1}(A))$, $A \in \mathcal{A}$ on measurable space $(\Omega, \mathcal{A})$. Thus, the function $f$ leads to a new function $\hat{f} : \Pi_{i \in N} \mathcal{P}_i \to \mathcal{P}$, where $\mathcal{P}_i$ is a set of strategies of an $i$-decision maker, $\mathcal{P}$ is a set of probability measures of outcomes.

We shall consider the situation from the point of view of an $i$-decision maker. Let us assume that a partial order of preference is defined on the set $\Omega$ according to the scheme described in the second case. As well as in the third case, there is some risk even if $|\mathcal{P}| = 1$. It also refers to the risk of unfavorable environment state (set of choices made by a community $N \setminus \{i\}$).

The case with $|\mathcal{P}| > 1$ is particularly interesting. Let us introduce the notation for a function which confronts the choice of the probability measure by an $i$-decision maker and the probability measure on the outcome space if the choices of other community members are fixed $\hat{f}[P_{\hat{i}}](P_i) = \hat{f}(P_{\hat{i}}, P_i)$, $P_i \in \mathcal{P}_i$, $P_{\hat{i}} \in \Pi_{j \in N \setminus \{i\}} \mathcal{P}_j$, $i \in N$. If such a partial order is defined for one $i$-decision maker on $\mathcal{P}$, that $\mathcal{P}$ and all its subsets of $\mathrm{Cod}(\hat{f}[P_{\hat{i}}])$ type, $P_{\hat{i}} \in \Pi_{j \in N \setminus \{i\}} \mathcal{P}$ are upper semilattices, then we have a problem of the game theory with the nature (with a randomized strategy of an $i$-decision maker) aimed at the search of such measure $P_i^*$ on $(\Delta_i, \mathcal{D}_i)$ that $\hat{f}(P_{\hat{i}}, P_i^*) = \sup \mathrm{Cod}(\hat{f}[P_{\hat{i}}])$ for any states of the nature $P_{\hat{i}} \in \Pi_{j \in N \setminus \{i\}} \mathcal{P}_j$. Here, as well as in the third case, we can speak about the risk of a nonoptimal choice. And finally, if a partial order of such kind is defined on $\mathcal{P}$ at least for one more decision maker apart from an $i$-decision maker, then we have a game theory problem in mixed strategies. Here, we can speak about both the risk of unfavorable choices made by a community $N \setminus \{i\}$ and the risk when an $i$-decision maker violates some principles of optimal solutions of game theory problems in mixed strategies.

The risks described in cases 3 and 4 are commonly referred to as risks in conditions of probabilistic (or scholastic) uncertainty.

In conclusion, it should be noticed that we have provided a classification of risk situations of systems development with their reference to the branches of mathematics which study them.

## 4 An example of a game theory problem which uses the relational risk concept

The proposed within this work relational risk concept can be difficult for perception. First of all, it focuses on the necessity to provide some minimally diverse algebraic structure assigned on the set of outcomes of system development, which let us speak about some risk. This structure is a system of semilattices nested in a certain way. In this case, a usual total order on the set of outcomes is a particular case of the structure under discussion. To facilitate the perception of such a risk concept in general and to demonstrate the constructivism of the concept, we shall turn to a simple and substantial example, though artificial to some extent.

We shall consider a situation of confrontation between two technologically different superpowers. The first one is considered to be a technological leader while the other one is an outsider. Let us assume that the superpowers are in the process of development of top priority scientific programs which are important for their political image. Let us present this situation in the form of a game [12]:

$$G = (Z_1, Z_2, u_1, u_2), \qquad (1)$$

where $Z_1$ is a set of possible strategies of the first player, $Z_2$ is a set of possible strategies of the second player, $u_1 : Z_1 \times Z_2 \to U_1$ is a gain function of the first player, $u_2 : Z_1 \times Z_2 \to U_2$ is a gain function of the second player, $U_1$ is a set of game outcomes for the first player, $U_2$ is a set of game outcomes for the second player.

Let us assume that $Z_1 = \{1, 2, 3\}$, $Z_2 = \{1, 2, 3\}$. Here, «1» means that the country does not have ambitious scientific programs, «2» means the implementation of the project of manned flight to Mars within the international cooperation framework, «3» means the project of manned flight to Mars implemented by one country. In this case, if both superpowers choose the strategy of cooperation, it is possible that there might be other countries participating in the project but not included into the formal description of a game, whereas the choice of strategy «2» by only one participant of a game means the implementation of the project in cooperation with some other countries except for the second player.

It should be noticed that the scale of the project under consideration cannot provide any evaluation of political, economic and scientific consequences of its implementation expressed in any numerical characteristics of the result. It is only possible to provide some qualitative comparison of the outcomes (one outcome is not better than the other). Although, even this is not always possible. Thus, gain functions $u_1$ and $u_2$ can be represented as partial orders of preference on the set

$$\left\{ \overline{z_1 z_2} : z_1 \in Z_1, z_2 \in Z_2 \right\}, \qquad (2)$$

where $\overline{z_1 z_2}$ means a term consisting of values $z_1$ and $z_2$. The term $\overline{z_1 z_2}$ means the result of the strategy choice made by the first superpower $z_1$, and $z_2$ is the strategy choice made by the second superpower.

Let us assume that the specialists in the field of scientific, technological and political expertise have provided the following orders for the first (see Fig. 4) and the second (see Fig. 5) players. While forming these orders, huge expenses spent for the implementation of the project (projects), success/ failure forecasts of the mission, social and economic consequences of the implementation and analyzing the project (projects) results, and political consequences have been taken into account.

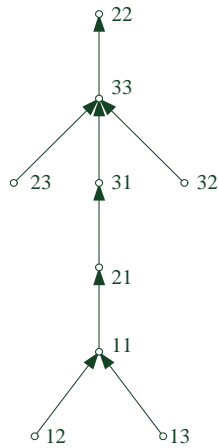
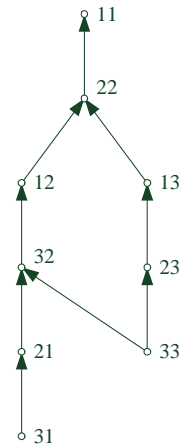

**Fig. 4.** A Hasse diagram of preference describing a gain function of the first player

**Fig. 5.** A Hasse diagram of preference describing a gain function of the second player

As you can see, the partial orders in Fig. 4 and 5 assign two different upper semi-lattices on the set (2). Besides, the following ratios are correct:

$$\text{Cod}(u_1[z_2=1]) = \left\{ \begin{array}{c} 31 \\ \uparrow \\ 21 \\ \uparrow \\ 11 \end{array} \right\}, \qquad (3)$$

$$\text{Cod}(u_1[z_2 = 2]) = \left\{ \begin{array}{c} 22 \\ \diagup \diagdown \\ 12 \quad 32 \end{array} \right\}, \tag{4}$$

$$\text{Cod}(u_1[z_2 = 3]) = \left\{ \begin{array}{c} 33 \\ \diagup \diagdown \\ 13 \quad 23 \end{array} \right\}, \tag{5}$$

$$\text{Cod}(u_2[z_1 = 1]) = \left\{ \begin{array}{c} 11 \\ \diagup \diagdown \\ 12 \quad 13 \end{array} \right\}, \tag{6}$$

$$\text{Cod}(u_2[z_1 = 2]) = \left\{ \begin{array}{c} 22 \\ \diagup \diagdown \\ 21 \quad 23 \end{array} \right\}, \tag{7}$$

$$\text{Cod}(u_2[z_1 = 3]) = \left\{ \begin{array}{c} 32 \\ \diagup \diagdown \\ 31 \quad 33 \end{array} \right\}, \tag{8}$$

In ratios (3)-(8), the preferences for the elements of the corresponding sets are shown graphically. They are induced by the orders given in Fig. 4 and 5. These preferences are also seen to assign the structures of the upper semilattices on sets (3)-(8). Thus, in game (1) the conditions of the second variant given in the previous section are implemented for both players, i.e. for each player there is a risk of nonoptimal (in the assigned sense) decision.

Let us consider what optimality principles can be used when the decisions are made by each player in the situation described in game (1). We shall compare the outcomes when a player chooses a strategy with all the possible strategies chosen by the other player (see Tab. 1). A question mark in the table denotes the cases when the outcomes, according to the experts, cannot be compared with each other.

The comparisons given in Tab. 1 show that in the game under consideration both the first and the second players do not have dominated or dominant strategies, i.e. all the strategies are undominated. In other words, in this game the principle of selection of dominant and elimination of dominated strategies is not applicable.

**Table 1.** Comparison of all the strategy pairs for each player

| Comparison of strategies of the first player (according to the order given in Fig. 1.1) | | | Comparison of strategies of the second player (according to the order given in Fig. 1.2) | | |
|---|---|---|---|---|---|
| 1 and 2 | 1 and 3 | 2 and 3 | 1 and 2 | 1 and 3 | 2 and 3 |
| 11 ≤ 21 | 11 ≤ 31 | 21 ≤ 31 | 11 ≥ 12 | 11 ≥ 13 | 12 ? 13 |
| 12 ≤ 22 | 12 ? 32 | 22 ≥ 32 | 21 ≤ 22 | 21 ? 23 | 22 ≥ 23 |
| 13 ? 23 | 13 ≤ 33 | 23 ≤ 33 | 31 ≤ 32 | 31 ? 33 | 32 ≥ 33 |

We shall try to explain the fact that the players have cautious strategies, i.e. strategies which can provide maximum gain for a player under a very unfavorable choice of the strategy made by the other player. As it can be seen in Fig. 4, for the first player $\inf_{z_2 \in Z_2} u_1(z_1, z_2)$ does not exist for all $z_1 \in Z_1$, consequently, $\sup_{z_1 \in Z_1} \inf_{z_2 \in Z_2} u_1(z_1, z_2)$ does not exist, which leads to the fact that the first player does not have any cautious strategies:

$$P_1(u_1) = \varnothing.$$

As you can see in Fig. 5, the following ratio is correct for the second player:

$$\sup_{z_2 \in Z_2} \inf_{z_1 \in Z_1} u_2(z_1, z_2) = u_2(3, 2).$$

i.e. the set of cautious strategies of the second player consists of one strategy:

$$P_2(u_2) = \{2\}.$$

This is the first optimality principle which can be used by the second player. The principle used to choose cautious strategies is particularly efficient when a player does not have enough information about the gain functions of the other players. Thus, he cannot forecast their rational behavior. In this case, the risk to violate this principle is a risk to obtain the worst outcome in two cases when the first player chooses strategies from set $\{2, 3\}$, see ratios (7) and (8). In case if the first player chooses strategy 1, then the cautious choice of the second player cannot be considered as the best one, see ratio (6), but it is a specific case which is acceptable as a part of precautionary behavior.

Let us assume now that the players are aware of their own and their opponent's gain functions. Which rational behavior might they choose in this case? They can provide a graph of the best answers for their opponent's strategies for each player, find the point where they meet, if there is one, and use their strategies which correspond to the points of intersection as the optimal ones expecting that the other players will do the same thing. None of the players will benefit if they reject following their strategies, as these strategies are considered to be the best answers to the strategies of the others. If one of them steps aside, it does not mean that the others will do the same, especially if the other player will benefit from this. The situation is referred to as the Nash equilibrium [12] and can be used as an optimality principle in our example. Let us draw the graphs of the players' best answers for game (1).

In our case, the graphs of the best answers of the first $BR_1(u_1)$ and the second $BR_2(u_2)$ players can be conveniently represented as:

$$\overline{z_1 z_2} \in BR_1(u_1) \underset{Def}{\Leftrightarrow} u_1(z_1, z_2) = \sup_{z \in Z_1} u_1(z, z_2),$$

$$\overline{z_1 z_2} \in BR_2(u_2) \underset{Def}{\Leftrightarrow} u_2(z_1, z_2) = \sup_{z \in Z_2} u_2(z_1, z).$$

It is easy to check that $BR_1(u_1) = \{31, 22, 33\}$, $BR_2(u_2) = \{11, 22, 32\}$. Using this result, we can calculate the set of the Nash equilibrium in game (1):

$$NE(G) = BR_1(u_1) \cap BR_2(u_2) = \{22\}.$$

Thus, according to the optimality principle under consideration, it is advisable for both players to choose strategy 2. Deviation from this equilibrium state is of no benefit for any of the players. The most favorable outcome for one of the players is possible only if the risk events for the second player are fully implemented including the computational errors and strategy choice.

It should be noticed that in the example given, we consider a risk under conditions of full certainty. Under conditions of complete awareness, a decision maker is able to theoretically calculate his/ her optimality strategy and keep to it. Here, risk relates not to the nature of the phenomena, but to the potential possibility of errors which might occur in the activity of a decision maker.

## 5    Results and discussion

The example considered demonstrates that it is possible to use partial orders assigned on the set of the game outcomes when making a decision. Herewith, traditional optimality principles of decisions such as dominated strategies elimination principle, dominant strategy equilibrium, cautious behavior, the Nash equilibrium, etc. can be interpreted in terms of partial orders. The application on these structures of corresponding minimal requirements with the reference to the order relation on the set of the game outcomes makes game situations easy-to-understand in terms of risk.

All in all, the interpretation of risk as a set of system development outcomes with preference order assigned on this set which forms a semilattice structure can be considered as a rather common one. If we do not take into consideration the orders on factor sets, then the only conclusion for the definition of risk can be a shift from the classical relation to the fuzzy relation. In other words, the order preference graph can be considered as a fuzzy relation. However, meaningfully, such an approach does not provide any new interpretations. Anyway, we can take risk into account only if a fuzzy relation defines the structure of a semilattice even, for example, on a significance threshold level. The only thing which is provided by the concept of fuzzy relation here is some parameterization capable of making experts' assessments better.

## 6     Conclusion

The basic result of this work is the formation of a new formal definition of risk. Thus, we can conclude that risk can be observed in a system when there is a preference order assigned to form a minimal structure of a semilattice on a bigger than just a one-point set of its development outcomes. Herewith, such an approach to risk interpretation is rather functional. There has been found a series of problem types which can be formulated in terms of the risk theory. Besides, the place of these types has been shown within a system of mathematical theories. A functional character of the relational risk theory developed has been illustrated with a rather substantial example.